\documentclass{bmvc2k}

\title{Future Semantic Segmentation with Convolutional LSTM}

\addauthor{Seyed shahabeddin Nabavi}{nabaviss@cs.umanitoba.ca}{1}
\addauthor{Mrigank Rochan}{mrochan@cs.umanitoba.ca}{2}
\addauthor{Yang Wang}{ywang@cs.umanitoba.ca}{3}

\addinstitution{
 Department of Computer Science
University of Manitoba\\
Winnipeg, MB, Canada\\
}

\runninghead{Nabavi, Rochan, Wang}{Future Semantic Segmentation}

\def\etal{\emph{et al}\bmvaOneDot}

\makeatletter
\DeclareRobustCommand{\cev}[1]{%
  \mathpalette\do@cev{#1}%
}
\newcommand{\do@cev}[2]{%
  \fix@cev{#1}{+}%
  \reflectbox{$\m@th#1\vec{\reflectbox{$\fix@cev{#1}{-}\m@th#1#2\fix@cev{#1}{+}$}}$}%
  \fix@cev{#1}{-}%
}
\newcommand{\fix@cev}[2]{%
  \ifx#1\displaystyle
    \mkern#23mu
  \else
    \ifx#1\textstyle
      \mkern#23mu
    \else
      \ifx#1\scriptstyle
        \mkern#22mu
      \else
        \mkern#22mu
      \fi
    \fi
  \fi
}

\makeatother

\usepackage{amssymb}

\begin{document}
\maketitle
\begin{abstract}
  We consider the problem of predicting semantic segmentation of future frames in a video. Given several observed frames in a video, our goal is to predict the semantic segmentation map of future frames that are not yet observed. A reliable solution to this problem is useful in many applications that require real-time decision making, such as autonomous driving. We propose a novel model that uses convolutional LSTM (ConvLSTM) to encode the spatiotemporal information of observed frames for future prediction. We also extend our model to use bidirectional ConvLSTM to capture temporal information in both directions. Our proposed approach outperforms other state-of-the-art methods on the benchmark dataset.
\end{abstract}

\section{Introduction}
\label{sec:intro}

\begin{figure}[t]
\begin{center}
  \includegraphics[width = \textwidth]{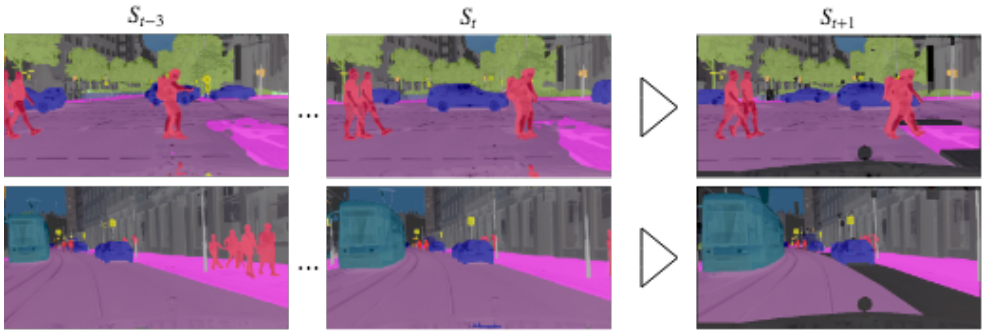}
\end{center}
   \caption{Illustration of future semantic segmentation. The first two columns show the input of the model. Given the semantic segmentation masks of several frames ($S_{t-3}...S_{t}$) in a video, our goal is to predict the semantic segmentation of an unobserved future frame $S_{t+1}$. }
\label{fig:problem}
\end{figure}

We consider the problem of future semantic segmentation in videos. Given several frames in a video, our goal is to predict the semantic segmentation of unobserved frames in the future. See Fig.~\ref{fig:problem} for an illustration of the problem. The ability to predict and anticipate the future plays a vital role in intelligent system decision-making \cite{sutton1998reinforcement,dosovitskiy2016learning}. An example is the autonomous driving scenario. If an autonomous vehicle can correctly anticipate the behaviors of other vehicles \cite{galceran2015multipolicy} or predict the next event that will happen in accordance with the current situation (e.g. collision prediction \cite{atev2005vision}), it can take appropriate actions to prevent damages.

Computer vision has made significant progress in the past few years. However, most standard computer vision tasks (e.g. object detection, semantic segmentation) focus on predicting labels on images that have been observed.  Predicting and anticipating the future is still challenging for current computer vision systems. Part of the challenge is due to the inherent uncertainty of this problem. Given one or more observed frames in a video, there are many possible events that can happen in the future. 

There has been a line of research on predicting raw RGB pixel values of future frames in a video sequence \cite{kalchbrenner2016video,mathieu2015deep,ranzato2014video,srivastava2015unsupervised}. While predicting raw RGB values of future frames is useful, it may not be completely necessary for downstream tasks. Another line of research focuses on using temporal correlation to improve current frame semantic segmentation stability \cite{li2018low,jin2017video,nilsson2016semantic,PatrauceanHC16}. In this paper, we focus on the problem of future semantic segmentation prediction~\cite{NextSegmPredICCV17}, where the goal is to predict the semantic segmentation of future frames. 

Future semantic segmentation is a relatively new problem in computer vision. There has been only limited work~\cite{NextSegmPredICCV17,jin2017predicting} on this topic. Luc~\etal~\cite{NextSegmPredICCV17} develop the first work on future semantic segmentation. Their model directly takes the segmentation masks of several frames as the input and produces the segmentation mask of a future frame. It does not explicitly captures the temporal relationship of the input frames. To address this limitation, Jin~\etal~\cite{jin2017predicting} propose a multi-task learning approach that jointly predicts optical flow and semantic segmentation of future frames. Since the optical flow captures the motion dynamics of adjacent frames, their approach implicitly models the temporal relationship of the input frames. The limitation of this approach is that optical flow estimation itself is a challenging task. In addition, it is more difficult to collect large scale dataset with ground-truth optical flow annotations. The method in \cite{jin2017predicting} uses the output of another optical flow estimation algorithm~(Epicflow \cite{revaud2015epicflow}) as the ground-truth. But this means the performance of this method is inherently limited by the performance of Epicflow.

In this paper, we propose a new approach for modeling temporal information of input frames in future semantic segmentation. Our method is based on convolutional LSTM, which has been shown to be effective in modeling temporal information \cite{NIPS2017_6689,finn2016unsupervised,PatrauceanHC16}. Unlike \cite{jin2017predicting}, our approach does not require the optical flow estimation. So our model is conceptually much simpler. Our model outperforms \cite{jin2017predicting} even though we do not use additional optical flow information.
    

%
%

In this paper, we make the following contributions. We propose a multi-level feature learning approach for future semantic segmentation. Our model uses convolutional LSTM (ConvLSTM) to capture the spatiotemporal information of input frames. We also extend ConvLSTM in our model to bidirectional ConvLSTM to further capture the spatiotemporal information from opposite directions. Our model outperforms the state-of-the-art approach in \cite{jin2017predicting} even without using the optical flow information.

\section{Related Work}
\label{sec:related}

In this section, we discuss several lines of research related to our work.

\noindent {\bf Future Prediction}: Recently, there is a line of research on future prediction in videos. Some of these works aim to predict the raw RGB values of future frames in a video. Ranzato \etal \cite{ranzato2014video} propose the first RNN/RCNN based model for unsupervised next frame prediction.  Srivastava \etal \cite{srivastava2015unsupervised} utilize LSTM \cite{hochreiter1997long} encoder-decoder to learn video representation and apply it in action classification. Villegas \etal \cite{villegas17mcnet} introduce a motion-content network to predict motion and content in two different encoders. Mathieu \etal \cite{mathieu2015deep} introduce a new loss function and a multi-scale architecture to address the problem of blurry outputs in future frame prediction. Vondrick \etal \cite{Vondrick_2016_CVPR} predict feature map of the last hidden layer of AlexNet \cite{krizhevsky2012imagenet} in order to train a network for anticipating objects and actions. Villegas~\etal~\cite{villegas2017learning} first estimate some high-level structure (e.g. human pose joints) in the input frames, then learn to evolve the high-level structure in future frames. There is also work~\cite{walker2015dense,PatrauceanHC16} on predicting future optical flows.

\noindent {\bf Future Semantic Segmentation Prediction}: Luc \etal. \cite{NextSegmPredICCV17} introduce the problem of future semantic segmentation prediction. They have introduced various baselines with different configurations for this problem. They have also considered several scenarios of future prediction, including short-term (i.e. single-frame), mid-term (0.5 second) and long term (10 seconds) predictions. An autoregressive method is designed to predict deeper into the future in their model. 

Jin et al.\cite{jin2017predicting} develop a multi-task learning framework for future semantic segmentation. Their network is designed to predict both optical flow and semantic segmentation simultaneously. The intuition is that these two prediction tasks can mutually benefit each other. Furthermore, they have introduced a new problem of predicting steering angle of vehicle as an application of semantic segmentation prediction in autonomous driving. However, their method requires ground-truth optical flow annotations, which are difficult to obtain.

\section{Approach}
\label{sec:approach}
In this section, we first present an overview of the proposed model in Sec. \ref{sec:overview}. We then describe our convolutional LSTM module in Sec. \ref{sec:ConvLSTM}. Finally, we introduce an extension of the ConvLSTM to bidirectional ConvLSTM in Sec.~\ref{sec:bilstm}.

\subsection{Model Overview} \label{sec:overview}
\begin{figure}[t]
\begin{center}
  \includegraphics[width = \textwidth]{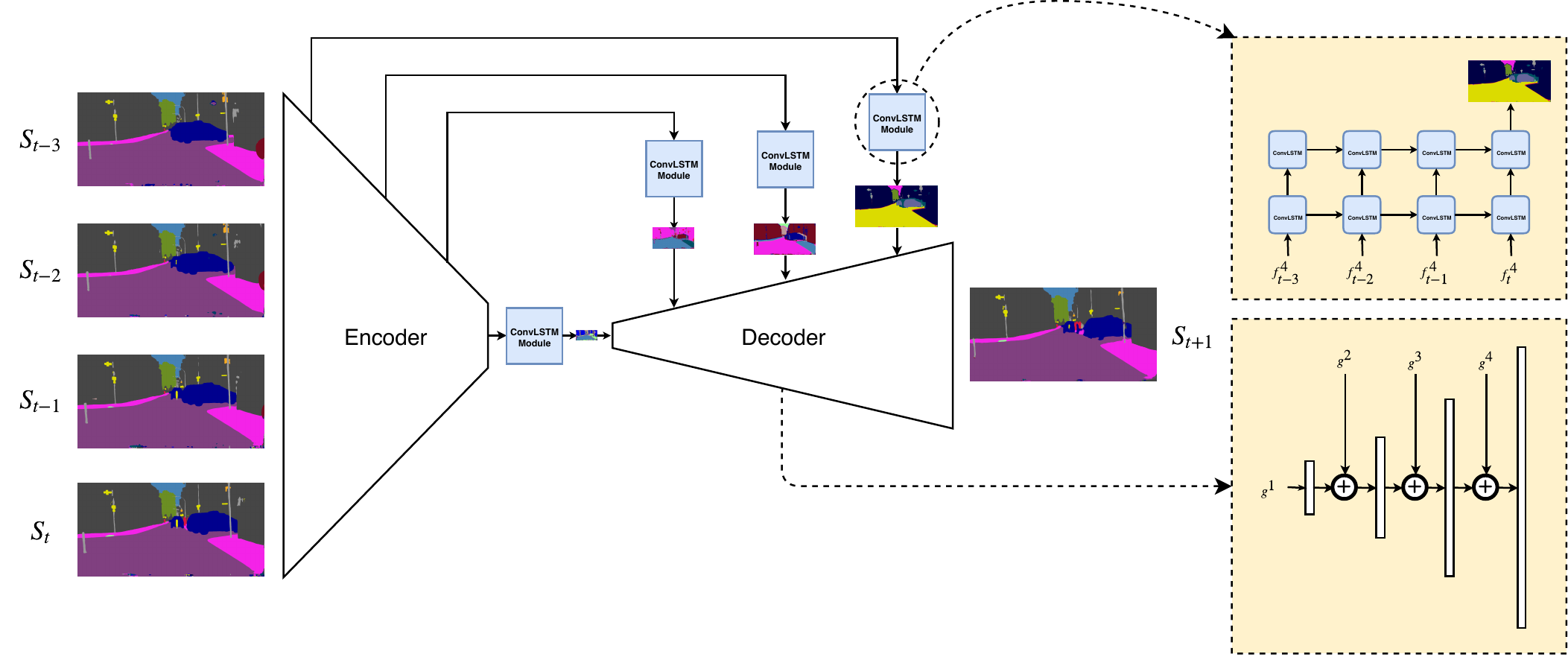}
\end{center}
   \caption{Overview of our proposed network for predicting scene parsing for one time ahead. Our network takes segmentation map ($S$) of video frames at $t-3$, $t-2$, $t-1$, and $t$ as an input and generates the segmentation map of the future frame $t+1$ as an output. The network consists of three major components: an encoder, convolutional LSTM (ConvLSTM) modules and a decoder. The encoder produces feature maps ($f^{k}_{t-3}:f^{k}_{t}$) for the inputs which are exploited by the ConvLSTM modules to predict the feature maps of future frame ($g^{k}$). Finally, the decoder which mainly has several deconvolution layers combines the outputs of different ConvLSTM modules and generate the segmentation map for the next time-step.}
\label{fig:Architecture-overall}
\end{figure}

Figure \ref{fig:Architecture-overall} shows the overall architecture of our proposed model. Our proposed network consists of three main components: an encoder, four convolutional LSTM (ConvLSTM) modules and a decoder. The encoder takes the segmentation maps of four consecutive frames at time $(t, t-1, t-2, t-3)$ and produce multi-scale feature maps for each frame. Each ConvLSTM module takes the feature map at a specific scale from these four frames as its input and captures the spatiotemporal information of these four frames. The outputs of these four ConvLSTM modules are then used by the decoder to predict the segmentation map of a future frame (e.g. at time $t+1$). In the following, we describe the details of these components in our model.

The encoder takes the semantic segmentation map of an observed frame and produces multi-scale feature maps of this frame. Following previous work \cite{jin2017predicting}, we use ResNet-101 \cite{he2016deep} as the backbone architecture of the encoder. We replace the last three convolution layers of ResNet-101 with dilated convolutions of size $2\times 2$ to enlarge the receptive field. We also remove the fully-connected layers in ResNet-101. In the end, the encoder produces multi-scale feature maps on each frame. Features at four different layers (``conv1'', ``pool1'',``conv3-3'',``conv5-3'') in the feature maps are then used as inputs to the four ConvLSTM modules. Let $(S_{t},S_{t-1},S_{t-2}, S_{t-3})$ be the semantic segmentation maps of the frames at time $(t, t-1, t-2, t-3)$, we use $(f_{t}^k, f_{t-1}^k, f_{t-2}^k, f_{t-3}^k)$ (where $k=1,2,3,4$) to denote the feature maps at the $k$-th layer for $(S_{t}, S_{t-1}, S_{t-2}, S_{t-3})$. In other words, $f_{t}^1$ will be the feature map at the ``conv1'' layer of the encoder network when using $S_{t}$ as the input. The spatial dimensions of $(f_{t}^k, f_{t-1}^k, f_{t-2}^k, f_{t-3}^k)$ are $(480\times480, 240\times240, 120\times120, 60\times60)$ when the input has a spatial size of $960\times 960$.

The $k$-th ($k=1,2,3,4$) ConvLSTM module will take the feature maps $(f_{t}^k,f_{t-1}^k,f_{t-2}^k,f_{t-3}^k)$ as its input. This ConvLSTM module produces an output feature map (denoted as $g^{k}$) which captures the spatiotemporal information of these four frames. 

We can summarize these operations as follows: 

\begin{equation}\label{eq:overview}
\begin{aligned}
& (f^{k}_{t},f^{k}_{t-1},f^{k}_{t-2},f^{k}_{t-3}) = Encoder^{k}(S_{t}, S_{t-1}, S_{t-2}, S_{t-3}) \hspace{0.2cm} \text{where $k$ =1,...,4}\\
& g^{k} = ConvLSTM^{k}(f^{k}_{t},f^{k}_{t-1},f^{k}_{t-2},f^{k}_{t-3}) \hspace{0.2cm} \text{ where $k$ = 1,...,4}\\
\end{aligned}
\end{equation}

Finally, the decoder takes the outputs ($g^{1}, g^{2}, g^{3}, g^{4}$) of the four ConvLSTM modules and produces the future semantic segmentation mask $S_{t+1}$ for time $t+1$ (assuming one-step ahead prediction). The decoder works as follows. First, we apply $1\times 1$ convolution followed by upsampling on $g^{1}$ to match the spatial and channel dimensions of $g^{2}$. The result is then combined with $g^{2}$ by an element-wise addition. The same sequence of operations ($1\times 1$ convolution,upsampling, element-wise addition) is subsequently applied on $g^{3}$ and $g^{4}$. Finally, another $1\times 1$ convolution (followed by upsampling) is applied to obtain $S_{t+1}$. These operations can be summarized as follows:
\begin{equation}
  \begin{aligned}
  & z^{1}=g^{1}, \ \  z^{k}=Up(C_{1\times 1}(z^{k-1}))+g^{k}, \textrm{ where } k=2,3,4\\
    & S_{t+1}=Up(C_{1\times 1}(z^{4}))
    \end{aligned}
\end{equation}
where $C_{1\times 1}(\cdot)$ and $Up(\cdot)$ denote $1\times 1$ convolution and upsampling operations, respectively.

\subsection{ConvLSTM module} \label{sec:ConvLSTM}
ConvLSTM is a powerful tool for capturing the spatiotemporal relationship in data \cite{NIPS2015_5955}, which is essential to predict the segmentation map of a future frame. We exploit this characteristic of ConvLSTM and introduce ConvLSTM modules at various stages in the model. In contrast to the conventional LSTMs that use fully connected layers in the input-to-state and state-to-state transitions, ConvLSTM uses convolutional layers instead. As shown in Fig. \ref{fig:Architecture-overall} (left), we have four ConvLSTM modules in our proposed network. The feature map from a specific layer in the encoder network (denoted as $f^{k}_{t-3}:f^{k}_{t}$ in Eq. \ref{eq:overview}) are used as the input to a ConvLSTM module. We set the kernel size to $3\times 3$ for convolutional layers in $ConvLSTM^{1}$ to $ConvLSTM^{3}$, whereas the $ConvLSTM^{4}$ has convolution with kernel size of $1\times 1$. Since the feature map of the future frame is based on the previous four consecutive video frames, the ConvLSTM unit has four time steps. The output of each ConvLSTM module is a feature map that captures the spatiotemporal information of the four input frames at a particular resolution.

Figure~\ref{fig:Architecture-overall} (right) shows the $k$-th ConvLSTM module. Each of the four input frames (at time $t-3, t-2, t-1, t$) corresponds to a time step in the ConvLSTM module. So the ConvLSTM module contains four time steps. We use $s$ to denote the time step in ConvLSTM module, i.e. $s\in\{t-3,t-2,t-1,t\}$. All inputs $f^k_{s}$, gates (input ($i_{s}$), output ($o_{s}$) and forget ($F_{s}$), hidden states $\mathcal{H}_{s}$, cell outputs  $\mathcal{C}_{s}$ are 3D tensors in which the last two dimensions are spatial dimensions. Eq. \ref{eq:ConvLSTM} shows the key operations of ConvLSTM:
\begin{equation}\label{eq:ConvLSTM}
\begin{aligned}
& i_{s} = \sigma(W_{fi}*f^k_{s} + W_{hi}*\mathcal{H}_{s-1} + W_{ci}  \circledast \mathcal{C}_{s-1} + b_{i}) \\
& F_{s} = \sigma(W_{fF}*f^k_{s} + W_{hF}*\mathcal{H}_{s-1} + W_{cF} \circledast \mathcal{C}_{s-1} + b_{F}) \\
& \mathcal{C}_{s} = F_{s} \circledast \mathcal{C}_{s-1} + i_{s} \circledast tanh(W_{fc} * f^k_{s} + W_{hc} * \mathcal{H}_{s-1} + b_{c}) \\
& o_{s} = \sigma(W_{fo} * f^k_{s} + W_{ho} * \mathcal{H}_{s-1} + W_{co} \circledast \mathcal{C}_{s} + b_{o})\\
& \mathcal{H}_{s} = o_{s} \circledast tanh(\mathcal{C}_{s}) \ \ \textrm { where } s=t-3, t-2, t-1, t\\
\end{aligned}
\end{equation}
\noindent where `*' denotes the convolution operation and `$\circledast$' indicates the Hadamard product. Since the desired output is the feature map of future frame $t+1$, we consider the last hidden state as the output of a ConvLSTM module, i.e. $g^{k}=\mathcal{H}_{t}$.

\subsection{ConvLSTM to Bidirectional ConvLSTM}\label{sec:bilstm}
Motivated by the recent success in speech recognition \cite{zhang2017very}, we further extend the ConvLSTM module to bidirectional ConvLSTM to model the spatiotemporal information using both forward and backward directions.  

\begin{figure}[t]
\begin{center}
  \includegraphics[scale=0.4]{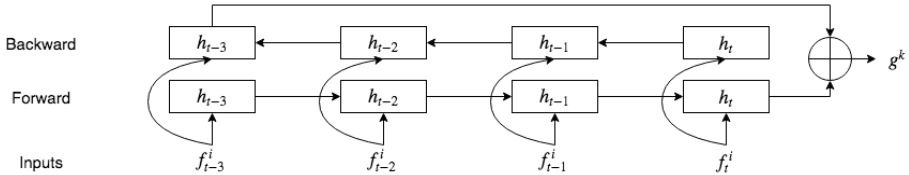}
\end{center}
   \caption{Architecture of bidirectional ConvLSTM module for future semantic segmentation.}
\label{fig:BCLSTM}
\end{figure}

Figure~\ref{fig:BCLSTM} illustrates the bidirectional ConvLSTM module that we propose for future semantic segmentation. Input feature maps $f^k_{t-3},...,f^k_{t}$ are fed to two ConvLSTM modules, $ConvLSTM^{forward}$ and $ConvLSTM^{backward}$. $ConvLSTM^{forward}$ computes the forward hidden sequence $\vec{\mathcal{H}}_{t+1}$ from time step $t-3$ to $t$, whereas $ConvLSTM^{backward}$ computes $\cev{\mathcal{H}}_{t+1}$ by iterating over inputs in the backward direction from time step $t$ to $t-3$. Finally, we concatenate the output of $ConvLSTM^{forward}$ and $ConvLSTM^{backward}$ and obtain feature map $g^k$ that is forwarded to the decoder for the subsequent processing. We can write these operations within bidirectional ConvLSTM as follows:

\begin{equation}\label{eq:BiConvLSTM}
\begin{aligned}
& \vec{\mathcal{H}}_{s},\vec{C}_{s} = ConvLSTM^{forward}(f^k_{s-1},\vec{\mathcal{H}}_{s-1},\vec{C}_{s-1})\\
  & \cev{\mathcal{H}}_{s},\cev{C}_{s} = ConvLSTM^{backward}(f^k_{s+1},\cev{\mathcal{H}}_{s+1},\cev{C}_{s+1}), \ \ \textrm { where } s=t-3, t-2, t-1, t\\
& g^k_{s} = concat(\vec{\mathcal{H}}_{t}, \cev{\mathcal{H}}_{t-3}) \\
\end{aligned}
\end{equation}

\section{Experiments}
\label{sec:exp}
In this section, we first discuss the dataset and experimental setup in Sec.~\ref{sec:setup}. We then present both quantitative and qualitative results in Sec.~\ref{sec:results}.

\subsection{Experimental Setup}\label{sec:setup}

\subsubsection{Datasets and evaluation metric}
We conduct our experiments on the Cityscapes dataset \cite{cordts2016cityscapes}. This dataset contains 2,975 training, 500 validation and 1,525 testing video sequences. Each video sequence has 30 frames and is 1.8 sec long. Every frame in a video sequence has a resolution of $1024\times2048$ pixels. Similar to previous work, we use 19 semantic classes of this dataset.

Following prior work \cite{NextSegmPredICCV17,jin2017predicting}, we evaluate the predicted segmentation maps of our method using the mean IoU (mIoU) on the validation set of the Cityscapes dataset.

\subsubsection{Baselines}
To demonstrate the effectiveness of our proposed model, we compare the performance of our model with the following baseline methods:

i) Jin \etal \cite{jin2017predicting}: The key component of this method is that it combines optical flow estimation and semantic segmentation in future frames. It uses the Res101-FCN architecture (a modified version of ResNet-101 \cite{he2016deep}) as the backbone network and the segmentation generator for the input. Since the code of \cite{jin2017predicting} is not publicly available, we have reimplemented the method in PyTorch. Note that Jin \etal \cite{jin2017predicting} report 75.2\% mIoU of Res101-FCN for the semantic segmentation task on the validation of Cityscapes dataset. But our re-implementation obtains only 71.85\% mIoU (see Table \ref{table:seg_performance}). However, our implementation of the PSPNet gives semantic segmentation performance similar to Res101-FCN reported in \cite{jin2017predicting}. 
 
ii) \textit{S2S} \cite{NextSegmPredICCV17}: This is one of state-of-the-art architecture for the future semantic segmentation. 

iii) \textit{Copy last input}: In this baseline, we copy the last input segmentation map ($S_{t}$) as the prediction at time $t+1$. The baseline is also used in  \cite{jin2017predicting}.

\begin{table}[h]
	\begin{center}
		\begin{tabular}{|l|c|}
			\hline
			Model & mIoU\\
			\hline\hline
			Res101-FCN \cite{jin2017predicting}  & {75.20} \\
			Res101-FCN \cite{jin2017predicting}* (our implementation) & {71.85} \\
			PSPNet \cite{zhao2017pyramid} & {75.72} \\
			\hline
		\end{tabular}
	\end{center}
	\caption{The performance (in terms of mIoU) of various backbone network architectures evaluated on the regular semantic segmentation task using the validation set of the Cityscapes dataset. *Performance of our implementation of Res101-FCN (2nd row) is lower than the original Res101-FCN reported in \cite{jin2017predicting} (1st row). But the performance of our PSPNet implementation (3rd row) is similar to Res101-FCN reported in \cite{jin2017predicting}.}
	\label{table:seg_performance}
\end{table}

\subsubsection{Implementation details}
We follow the implementation details of Jin \etal \cite{jin2017predicting} throughout our experiments. Similar to \cite{jin2017predicting}, we use Res101-FCN as the backbone architecture of our model. We set the length of the input sequence to 4 frames, i.e., segmentation maps of frames at $t-3, t-2, t-1$ and $t$ are fed as the input to predict the semantic segmentation map of the next frame $t+1$. For data augmentation, we use random crop size of $256\times256$ and also perform random rotation. Following prior work, we consider the 19 semantic classes in the Cityscapes dataset for prediction. We use the standard cross-entropy loss function as the learning objective. The network is trained for 30 epochs in each experiment which takes about two days using two Titan X GPUs.

\subsection{Results} \label{sec:results}

In this section, we present the quantitative performance of our model for future semantic segmentation and compare with other state-of-the-art approaches. Following prior work, we consider both one time-step ahead and three time-steps ahead predictions. We also present some qualitative results to demonstrate the effectiveness of our model.

Since the Cityscapes dataset is not fully annotated, we follow prior work~\cite{NextSegmPredICCV17,jin2017predicting} and use a standard semantic segmentation network to produce segmentation masks on this dataset and treat them as the ground-truth annotations. These generated ground-truth annotations are then used to learn the future semantic segmentation model.

\subsubsection{One time-step ahead prediction}
We first evaluate our method in one time-step ahead prediction. In this case, our goal is to predict the future semantic segmentation of the next frame. Table \ref{table:one_time_ahead} shows the performance of different methods on the one-time ahead semantic segmentation prediction.

Table \ref{table:one_time_ahead} shows the performance when the ground-truth semantic segmentation is generated by Res101-FCN (``Ours (Res101-FCN)'' in Table~\ref{table:one_time_ahead}) and PSPNet (``Our (PSPNet)'' in Table~\ref{table:one_time_ahead}). Note that the backbone architecture of our model is Res101-FCN in either case. The two sets of results (``Ours (Res101-FCN)'' and ``Our (PSPNet)'') only differ in how the ground-truth semantic segmentation used in training is generated. The Res101-FCN network identical to \cite{jin2017predicting} is used as the backbone architecture of our model in both cases. 

We also compare with other state-of-the-art approaches in Table~\ref{table:one_time_ahead}. It is clear from the results that our method using ConvLSTM modules significantly improves the performance over the state-of-the-art. When we use bidirectional ConvLSTM modules in our model, we see further improvement in the performance (nearly 5 \%). In addition, we also compare the performance of a baseline method where we simply remove the ConvLSTM modules (i.e. Ours (w/o ConvLSTM)) from the proposed network. Instead, we concatenate the feature maps $(f_{t}^k, f_{t-1}^k, f_{t-2}^k, f_{t-3}^k)$ after corresponding $1\times 1$ convolution and upsampling to make their dimensions match. Then we apply a simple convolution on the concatenated feature maps to produce $g^{k}$. These results demonstrate the effectiveness of the ConvLSTM modules for the future semantic segmentation task.

\begin{table}[!htb]
    \begin{center}
        \begin{tabular}{|l|c|}
           \hline
           Model & mIoU\\
           \hline\hline
            S2S \cite{NextSegmPredICCV17}  & {62.60\textsuperscript{\ddag}} \\ 
           Jin \etal \cite{jin2017predicting} & {66.10}\\ 
           Copy last input  & {62.65}\\ 
           \hline
           Ours (Res101-FCN) & \\
           \quad \quad w/o ConvLSTM & {60.80}*\\ 
           \quad \quad ConvLSTM & {64.82}*\\ 
           \quad \quad Bidirectional ConvLSTM & {65.50}*\\
           \hline
           Ours (PSPNet) & \\
           \quad \quad w/o ConvLSTM & {67.42}\\ 
           \quad \quad ConvLSTM & {\bf 70.24}\\ 
           \quad \quad Bidirectional ConvLSTM & {\bf 71.37 }\\
           \hline
        \end{tabular}
    \end{center}
    \caption{The performance of future semantic segmentation on the validation set of the Cityscapes dataset for one time-step prediction. We show the results of using both Res101-FCN and PSPNet for generating the ground-truth semantic segmentation. * indicate that input sequence is generated using our implementation of Res101-FCN (see Table\ref{table:seg_performance}). \textsuperscript{\ddag}Results taken from Jin \etal \cite{jin2017predicting}.}
    \label{table:one_time_ahead}
\end{table}

\subsubsection{Three time-steps ahead prediction}
Following Luc \etal \cite{NextSegmPredICCV17}, we also evaluate the performance of our model in a much more challenging scenario. In this case, the goal is to predict the segmentation map of the frame that is three time-steps ahead. Table \ref{table:three_time_ahead} shows the performance of different methods on this task. For the results in Table~\ref{table:three_time_ahead}, we have used PSPNet to generate the ground-truth semantic segmentation. It is clear from the results that our method with ConvLSTM modules performs very competitively. When we bidirectional ConvLSTM modules in our model, the performance is further improved. In particular, our method with bidirectional ConvLSTM achieves the state-of-the-art performance. Again, we also compare with the baseline ``Ours (w/o ConvLSTM)''. These results demonstrate the effectiveness of the ConvLSTM modules for the future semantic segmentation task.

\begin{table}[!htb]
    \begin{center}
      \centering
        \begin{tabular}{|l|c|}
\hline
Model & mIoU\\
\hline\hline
S2S(GT) \cite{NextSegmPredICCV17}  & {59.40} \\ 
Copy last input  & {51.08}\\ 
\hline
Ours (w/o ConvLSTM) & {53.70}\\ 
Ours (ConvLSTM) & {58.90}\\ 
Ours (Bidirectional ConvLSTM) & {\bf 60.06 }\\
\hline
        \end{tabular}
    \end{center}
    \caption{The performance of different methods for three time-steps ahead frame segmentation map prediction on the Cityscapes validation set. We show performance when using PSPNet to generate the ground-truth semantic segmentation.}
    \label{table:three_time_ahead}
\end{table}

\subsubsection{Qualitative results}
\begin{figure}
	\begin{center}
		\setlength\tabcolsep{1pt}
		\def\arraystretch{0.5}
		\begin{tabular}{cc@{\hskip 10pt}cc}
			\includegraphics[height=0.76in]{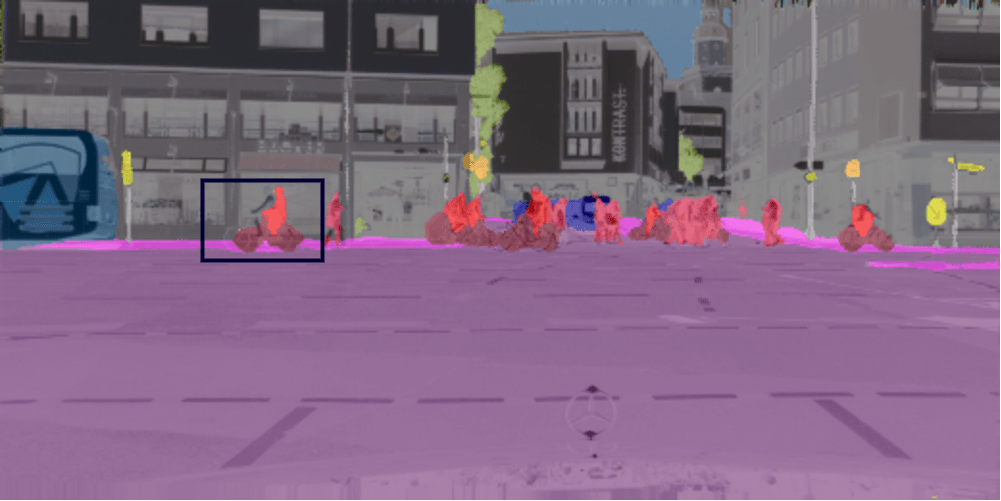}&
			\includegraphics[height=0.76in]{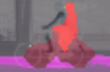}&   
			\includegraphics[height=0.76in]{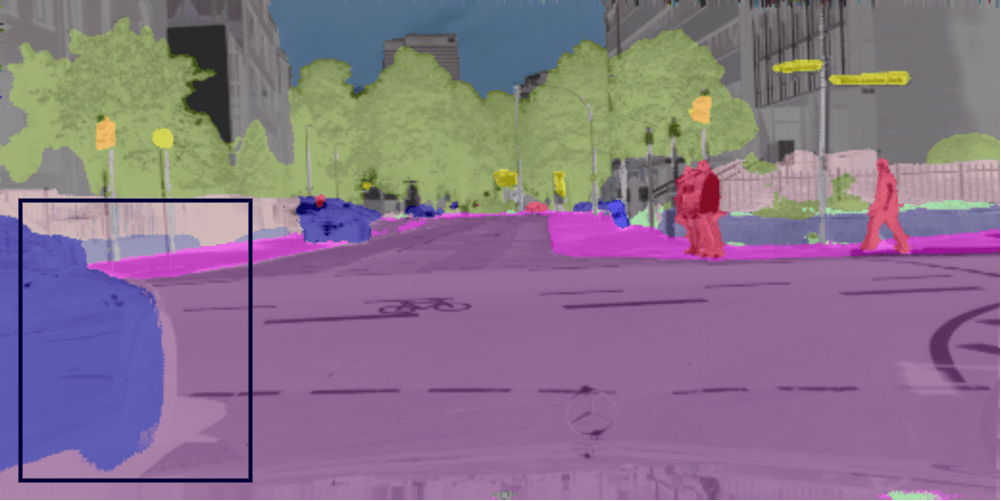}&
			\includegraphics[height=0.76in]{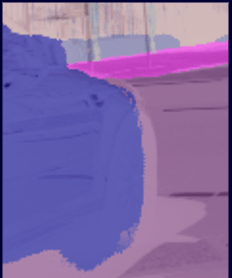}\\
			\includegraphics[height=0.76in]{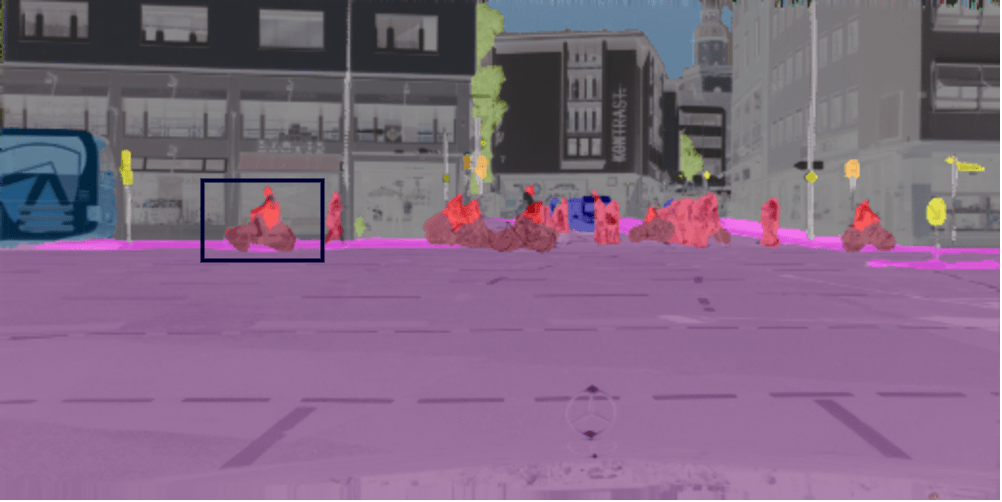}&
			\includegraphics[height=0.76in]{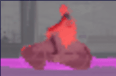}&   
			\includegraphics[height=0.76in]{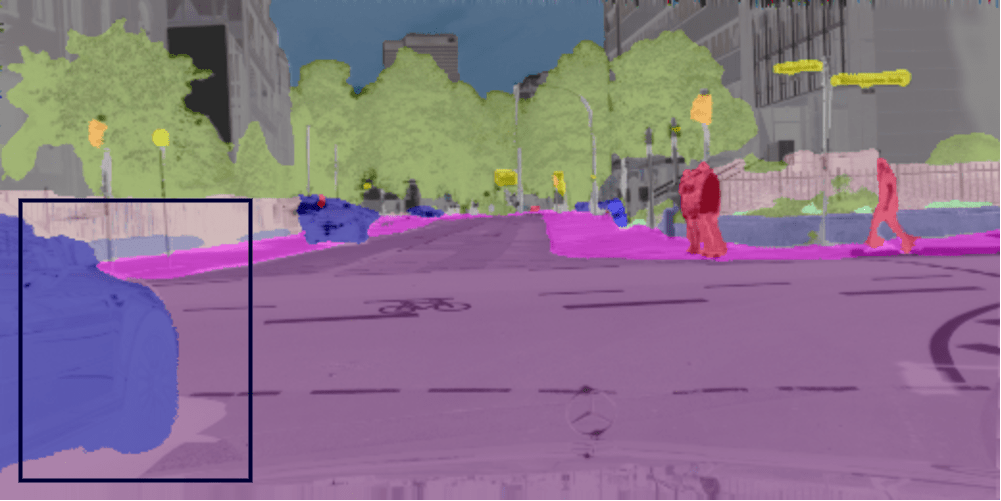}&
			\includegraphics[height=0.76in]{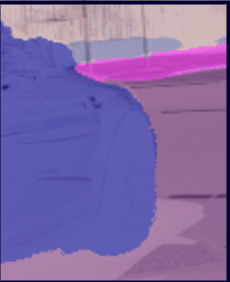}\\
			\includegraphics[height=0.76in]{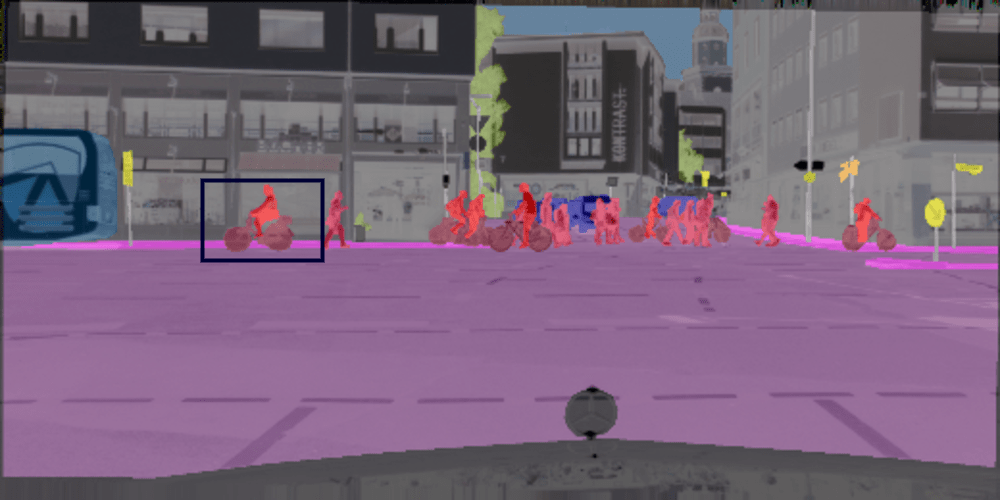}&
			\includegraphics[height=0.76in]{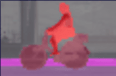}&   
			\includegraphics[height=0.76in]{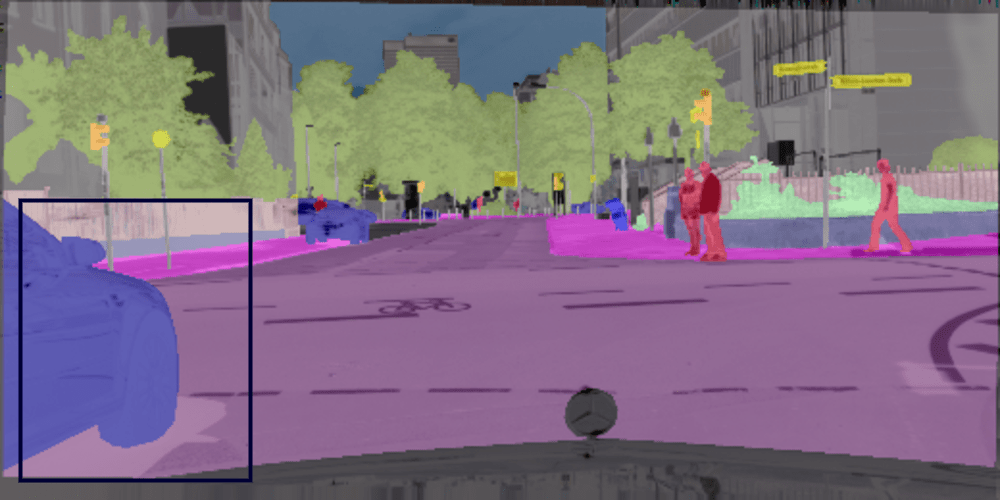}&
			\includegraphics[height=0.76in]{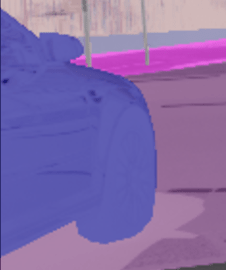}\\
		\end{tabular}
		\caption{Qualitative examples for the one time-step ahead prediction: (top) baseline Res101-FCN; (middle) our proposed model with ConvLSTM module; (bottom) ground truth. We show the segmentation mask on the entire image (1st and 3rd column) and the zoom-in view on a patch indicated by the bounding box (2nd and 4th column). This figure is best viewed in color with magnification.}
		\label{fig:qualitative-1}
	\end{center}
\end{figure}

\begin{figure}
	\begin{center}
		\setlength\tabcolsep{1pt}
		\def\arraystretch{0.5}
		\begin{tabular}{cc@{\hskip 10pt}cc} 
			\includegraphics[height=0.9in]{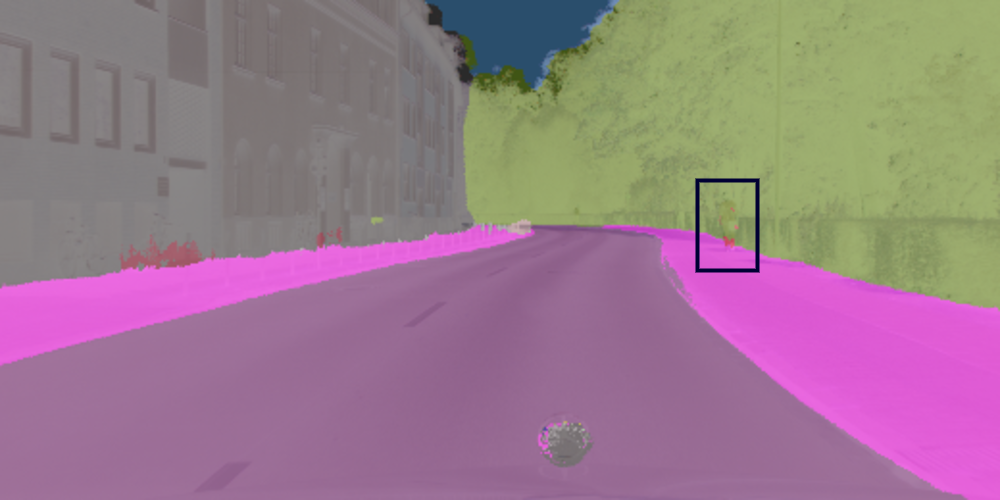}&
			\includegraphics[height=0.9in]{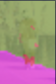}&   
			\includegraphics[height=0.9in]{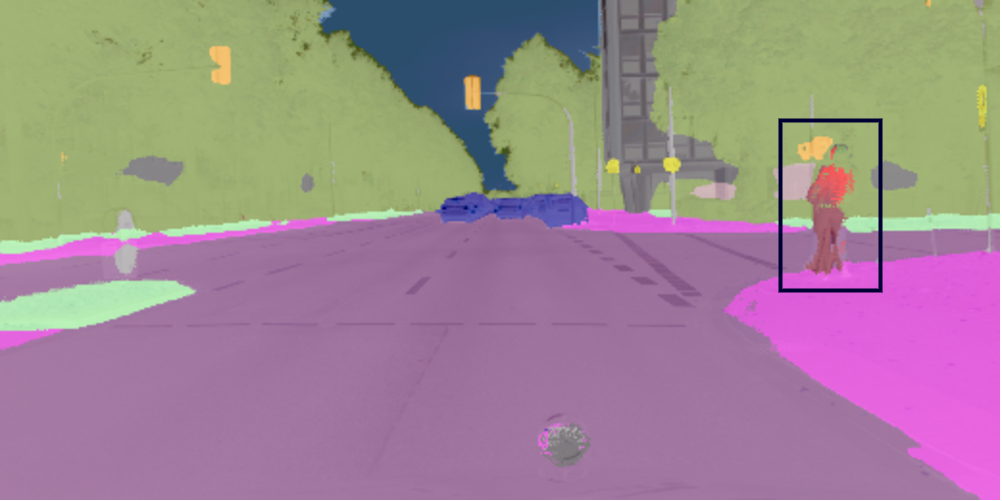}&
			\includegraphics[height=0.9in]{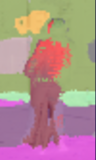}\\
			\includegraphics[height=0.9in]{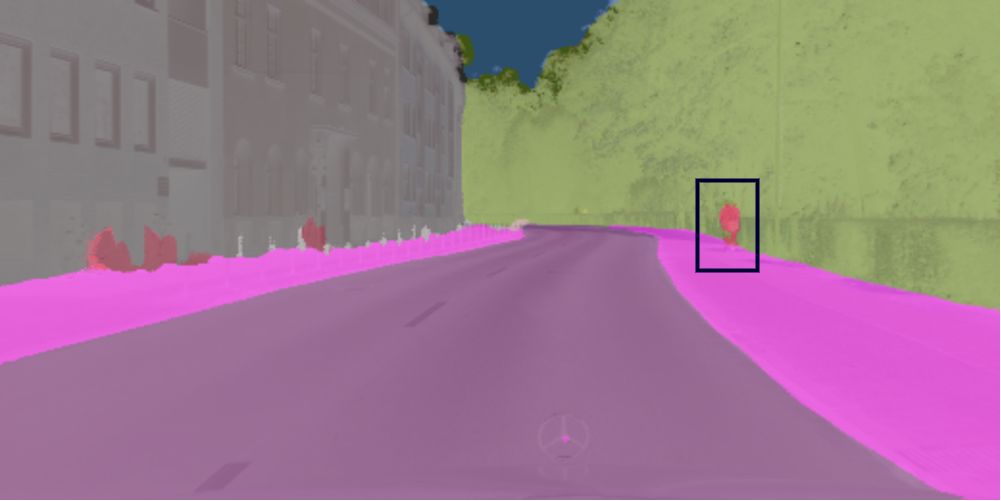}&
			\includegraphics[height=0.9in]{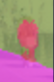}&   
			\includegraphics[height=0.9in]{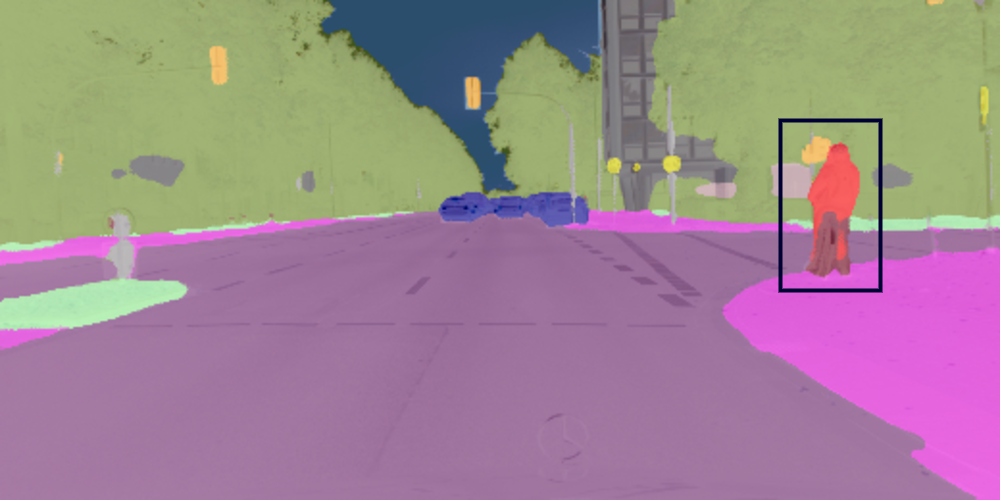}&
			\includegraphics[height=0.9in]{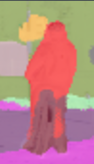}\\
			\includegraphics[height=0.9in]{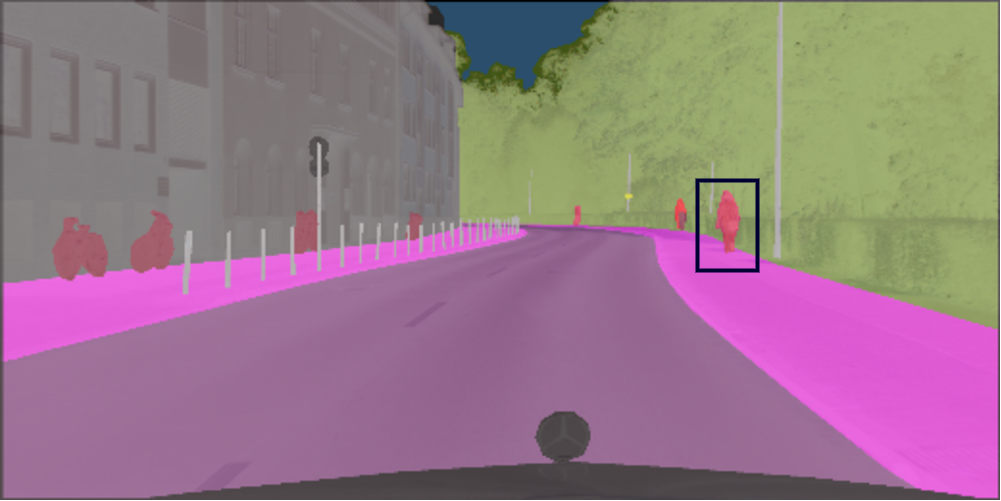}&
			\includegraphics[height=0.9in]{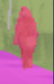}&   
			\includegraphics[height=0.9in]{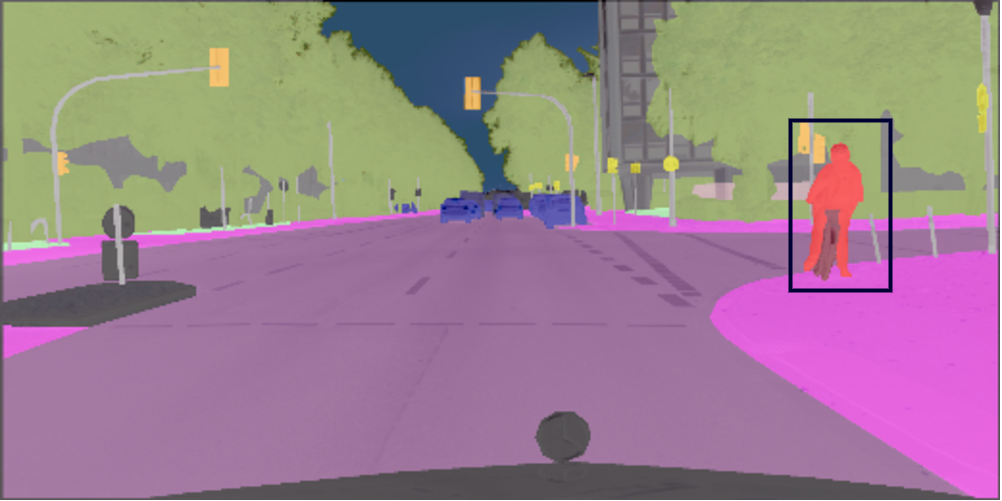}&
			\includegraphics[height=0.9in]{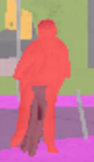}\\
		\end{tabular}
		\caption{Qualitative examples for the three time-steps ahead prediction: (top) baseline Res101-FCN; (middle) our proposed model with ConvLSTM module; (bottom) ground truth. We show the segmentation mask on the entire image (1st and 3rd column) and the zoom-in view on a patch indicated by the bounding box (2nd and 4th column). This figure is best viewed in color with magnification.}
		\label{fig:qualitative-2}
	\end{center}
\end{figure}

Figure \ref{fig:qualitative-1} shows examples of one time-step ahead prediction. Compared with the baseline, our model produces segmentation masks closer to the ground-truth. Figure \ref{fig:qualitative-2} shows examples of three time-steps ahead prediction, which is arguably a more challenging task. In this case, the improvement of our model over the baseline is even more significant.

\section{Conclusion}
\label{sec:concl}

We have introduced a new approach to predict the semantic segmentation of future frames in videos. Our approach uses the convolutional LSTM to encode the spatiotemporal information of observed frames. We have also proposed an extension using the bidirectional ConvLSTM. Our experimental results demonstrate that our proposed method significantly outperforms other state-of-the-art approaches in future semantic segmentation.

\section*{Acknowledgments}
This work was supported by a University of Manitoba Graduate Fellowship and a grant from NSERC. We thank NVIDIA for donating some of the GPUs used in this work.

\bibliography{egbib}
\end{document}